# Exploring PCA-based feature representations of image pixels via CNN to enhance food image segmentation

Ying Dai

**Abstract**

For open vocabulary recognition of ingredients in food images, segmenting the ingredients is a crucial step. This paper proposes a novel approach that explores PCA-based feature representations of image pixels using a convolutional neural network (CNN) to enhance segmentation. An internal clustering metric based on the silhouette score is defined to evaluate the clustering quality of various pixel-level feature representations generated by different feature maps derived from various CNN backbones. Using this metric, the paper explores optimal feature representation selection and suitable clustering methods for ingredient segmentation. Additionally, it is found that principal component (PC) maps derived from concatenations of backbone feature maps improve the clustering quality of pixel-level feature representations, resulting in stable segmentation outcomes. Notably, the number of selected eigenvalues can be used as the number of clusters to achieve good segmentation results. The proposed method performs well on the ingredient-labeled dataset FoodSeg103, achieving a mean Intersection over Union (mIoU) score of 0.5423, outperforming the state-of-the-art results. Importantly, the proposed method is unsupervised, and pixel-level feature representations from backbones are not fine-tuned on specific datasets. This demonstrates the flexibility, generalizability, and interpretability of the proposed method, while reducing the need for extensive labeled datasets.

## 1. Introduction

Segmenting different ingredients from food images and identifying them holds significant importance in both food safety and agriculture, affecting various aspects of food production, nutritional health. It fosters a deeper understanding of nutritional needs, promotes sustainable agricultural practices, and encourages innovation and collaboration across sectors. By leveraging the insights gained from ingredient segmentation, stakeholders can work towards healthier populations and more resilient agricultural systems. However, when applied to these fields, high accuracy and strong interpretability in food segmentation are strictly required.

Food segmentation belongs to the category of image segmentation. It is a challenge task when the shapes of objects and the background are complex. For example, a food image may contain multiple ingredients, which can appear in various shapes due to different cutting and cooking methods. The backgrounds are also complex due to various table and container placements.

Image segmentation has two different categories: two-stage approaches and one-stage approaches. The former decouples the problem into a class-agnostic segmentation task and a pixel-level mask classification task. The latter directly output each pixel-level mask and its label. However, an intuitive observation is that, when given an image for semantic segmentation, humans naturally group pixels into segments first and then assign semantic labels at the segment level. For instance, a child can easily group the pixels of an object even without knowing its name [1]. So, two-stage approaches explicitly focus on objects, leading to better recall compared to single-stage methods [3]. As a result, they are more commonly employed in open vocabulary object detection tasks [2]. Accordingly, this paper focuses on the two-stage approach and proposes a new method to enhance the ability to group image pixels with the similar semantic features into a same segment, thereby improving the performance of semantic segmentation. This is particularly relevant in the case study of segmenting different ingredients from food images.

The state-of-the-art (SOTA) research regarding two-stage image segmentation in the literature includes [3-7]. The authors in [3] introduced an image segmentation task which aimed to segment all visual entities (objects and stuffs) in an image without predicting their semantic labels. This paper proposed a CondInst[4]-like fully-convolutional architecture with two novel modules specifically designed to exploit the class-agnostic and non-overlapping requirements of entity segmentation. The authors in [5] proposed to decouple the zero-shot semantic segmentation (Z3S) into two subtasks and train a Z3S model in the two-stage approach. Authors in [6] introduced a segment anything model (SAM) which was designed and trained to be promptable. The authors in [7] proposed a mask classification model with predict a set of masks, each associated with a single global class label prediction. The two-stage method simplified the landscape of approaches to semantic segmentation tasks and showed good results. All these papers use the ResNet [12] backbones or vision transformer (ViT)-based Swin-Transformer [13] to obtain feature representations of image pixels for generating a set of masks. However, whether the ResNet or Swin-Transformer are the best backbones and whether there are other more effective backbones and methods to represent the pixels are not explored. Furthermore, the papers [3-6] used pixel-level manually-annotated masks to train the segmentation model with the mask loss. These segmentation methods heavily depend on the human-labeled data which is both labor-intensive and costly and thus less scalable [8]. The paper [7] utilized the pre-trained transformer as a decoder to obtain the mask predictions. However, whether the transformer is optimal as a decoder is unclear.

Due to the limitation of pixel-level supervised semantic segmentation, several recent works [8-11] emerged as promising approaches for unsupervised image segmentation. The paper [10] exploited self-supervised learning models like DINO [9] to 'discover' objects and train a detection and segmentation model using a cut-and-learn pipeline. The paper [11] harnessed the semantically rich feature correlations produced by unsupervised methods like DINO for segmentation with the method of k-nearest neighbors (KNNs). The authors in [8] vector quantized the model targets (pseudo semantic labels) by clustering the instance level features generated by [10] with k-means to train a segmentation model. However, all these papers rely on self-supervised ViT features and clustering methods like

KNNs or k-means to generate semantic masks. The potential of other models to provide more effective pixel-level feature representations, or alternative clustering methods to generate better semantic masks without additional training, remains unexplored.

In this paper, we aim to explore more effective and explainable pixel-level feature representation methods using backbones beyond ResNet and Swin Transformer. Additionally, we investigate clustering methods better suited for these pixel-level feature representation arrays to segment food images.

Our contributions are summarized as follows.

- We conduct empirical experiments to investigate the impact of different backbones and different feature representations on the clustering quality of pixels.
- We conduct a metric of assessing the clustering quality of data points, and use it to select the suitable backbones and pixel-level feature representations, and the appropriate clustering methods for generating pixel-wise masks to segment instances.
- We propose a novel method for representing pixel-wise features by utilizing the techniques of principle component analysis (PCA) and feature map concatenation.
- We propose a new method for predicting the number of clusters in clustering based on the eigenvalues of pixel-wise feature representation array.
- We validate the effectiveness of the proposed approach in multiple ingredient segmentation on public dataset.

## 2. Related works

### 2.1 Food segmentation

Food segmentation involves segmenting different ingredients of food items in images. This can include separating individual food items on a plate, distinguishing between different ingredients in a dish, or recognizing food packaging. Despite advancements, food segmentation research faces several challenges: variability in food appearance, occlusion of food items, diverse and clutter backgrounds, and lack of Annotated Datasets. In [18], authors propose a multi-modality pre-training approach called ReLeM that explicitly equips a segmentation model with rich and semantic food knowledge. In [19], authors present a Bayesian version of two different state-of-the-art semantic segmentation methods to perform multi-class segmentation of foods and estimate the uncertainty about the given predictions. In [20], authors propose a novel framework to integrates the coarse semantic mask with SAM-generated masks to enhance semantic segmentation quality. In [21], two models are trained for food segmentation, one based on convolutional neural networks and the other on Bidirectional Encoder representation for Image Transformers (BEiT). In [22], Authors introduce a framework that adopts an open-vocabulary setting and enhances text embeddings with visual context. However, all these approaches are class-based segmentation, and models are trained or fine-tuned on specific datasets with the pixel-wise annotations.

As food items are highly diverse and often unlabelable by strict classes, class-agnostic segmentation has become a valuable area of SOTA research. This approach focuses on segmenting items without requiring predefined classes, making it adaptable to new or unique food items, especially in open-set or open-vocabulary scenarios. In this paper, we propose a novel approach for class-agnostic segmentation, and don't need to retrain models on specific datasets with the pixel-wise annotations.

*2.2 Public dataset*

There are two datasets widely used in training and validating food segmentation models: FoodSeg103 [18] and UECFOODPIX COMPLETE [24]. FoodSeg103 contain 2,135 images for testing. These images are annotated with 103 ingredient names and each image has an average of 6 ingredient labels and pixel-wise masks. UECFOODPIX COMPLETE contain 1000 images for testing. These images are annotated with 102 either dish or ingredient names and each image has pixel-wise masks. In this paper, we utilize these two datasets to validate the proposed method.

## 3. Some observations

From beginning, we introduce an internal clustering metric that assesses the clustering quality of data points: silhouette rate (SR). SR is derived from silhouette score (SS). Silhouette score measures how similar an individual Point is to its own cluster (cohesion) compared to other clusters (separation). The score ranges from -1 to 1, where a higher score indicates the better-defined cluster. Accordingly, SR is defined by the following expression (1).

$$SR = \text{(the number of points with SS} > T\text{)}/\text{(the number of all points)} \quad (1)$$

It means the rate of points which have the SS larger than a threshold T to all points. A higher SR indicates better clustering quality for points which are labeled.

Based on the experimental results of multiple ingredient segmentation in food images, several issues are identified. One is that clustering qualities of pixel-wise feature representations derived from resNet50 and SAM is worse than those from efficientNetB0. The pixel-wise feature representations from resNet50 and efficientNetB0 are derived by the feature maps of their last convolutional layer, while those from SAM are obtained by the embeddings of it. An example about the t-SNE plots of these feature representations corresponding to the labelled masks are shown in Fig. 1.

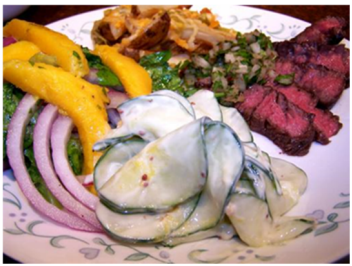 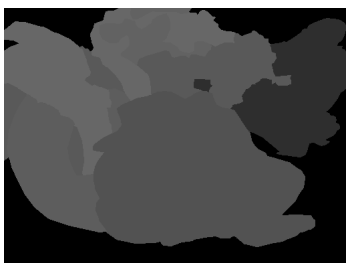

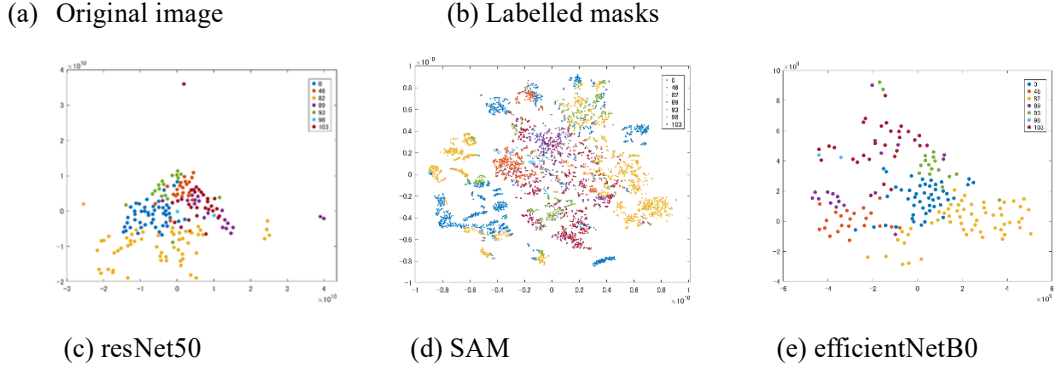

(a) Original image      (b) Labelled masks

(c) resNet50      (d) SAM      (e) efficientNetB0

Fig. 1 An example of three models' pixel-wise t-SNE plots corresponding to the labeled masks

Where, (a) is the original image and (b) is the labeled mask image, while (c), (d), and (e) are t-SNE plots of the pixel-wise feature representation array corresponding to the labeled masks. The blue points represent the distribution of background mask and the other color points represent that of each ingredient mask. Based on the t-SNE plots shown in Fig. 1, we observe that panel (e), representing the t-SNE of pixel-wise feature representations derived from EfficientNetB0, demonstrates the highest degree of inter-cluster separation and intra-cluster compactness. Specifically, the SR for EfficientNetB0 is 0.2969 when $T>0.3$, while the SR for ResNet50 is 0.0885 and is 0 for SAM. This finding clearly indicates that the clustering quality of EfficientNetB0 is significantly superior to that of ResNet50 and SAM.

Next, inspired by reports that the block-pruned EfficientNetB0 model can be used to improve the performance of ingredient segmentation [16], we further assess the clustering quality of different pixel-wise feature representation arrays derived from various models and generated through different methods. A summary of various feature representations is provided below. It is noted that the pruned EfficientNetB0 model used in this paper follows the configuration in [16], with blocks 6, 10, 13, and 14 removed. This model was subsequently fine-tuned on a single-ingredient image dataset comprising 110 categories.

feature representation (FR) 1: generated through the feature maps of head convolutional layer from efficietNetB0;

FR 2: generated through the top 5 principle component (PC) maps of the feature maps of head convolutional layer from efficietNetB0;

FR 3: generated through the concatenation of the feature amps of convolutional layers of head, block 15 and block 5 from efficietNetB0;

FR 4: generated through the top 5 PC maps of the concatenation of the feature maps of convolutional layers of head, block 15 and block 5 from efficietNetB0;

FR 5: generated through the feature map of head convolutional layer from pruned efficietNetB0;

FR 6: generated through the top 5 PC maps of the feature maps of head convolutional layer from pruned efficietNetB0;

FR 7: generated through the concatenation of the feature amps of convolutional layers of head, block 15 and block 5 from pruned efficietNetB0;

FR 8: generated through the top 5 PC maps of the concatenation of the feature maps of convolutional layers of head, block 15 and block 5 from pruned efficietNetB0.

Table 1 presents all the SR values of these feature representations for an image corresponding to the labeled masks.

Table 1 SR values

| FR | FR 1 | FR 2 | FR 3 | FR 4 | FR 5 | FR 6 | FR 7 | FR 8 |
|---|---|---|---|---|---|---|---|---|
| SR | 0.3 | 0.5781 | 0.276 | 0.6719 | 0.49 | 0.599 | 0.4323 | 0.6615 |

From the SR values in Table 1, we observe that FR 4 and FR 8 achieve the best results, followed by FR 2 and FR 6. This indicates that the pixel-wise FRs generated through principal components enhance their clustering quality. Compared to the corresponding FRs without principal component transformation, 75% of the improvements are greater than 0.2. For the pair of FR 3 and FR 4, FR 4 shows the highest SR value at 0.67, with the largest improvement rate of 0.58. Those suggests that the feature representations generated through the PC maps of the concatenated feature maps from efficientNetB0 and pruned efficientNetB0 provide excellent inter-cluster separation and intra-cluster compactness for the labeled pixels.

Fig. 2 shows the t-SNE plots of above FRs corresponding to the labeled masks.

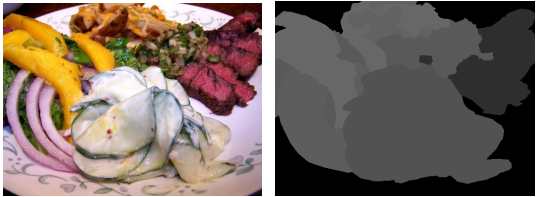

(a) Original image    (b) Labelled masks

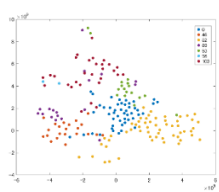 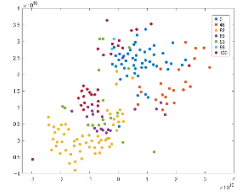 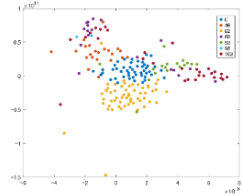 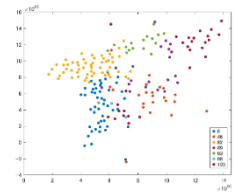

(c) FR 1                (d) FR 2                (e) FR 3                (f) FR 4

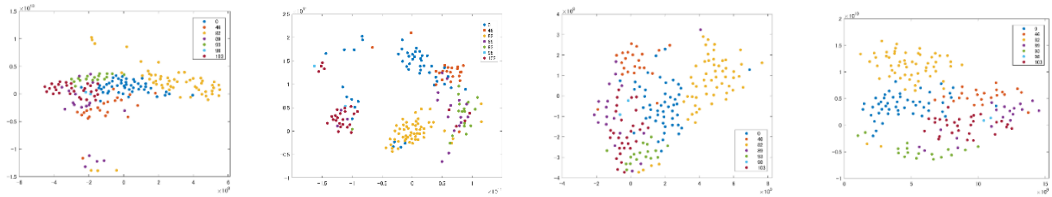

(g) FR 5              (h) FR 6              (i) FR 7              (j) FR 8

Fig. 2 An example of t-SNE plots of various pixel-wise FRs corresponding to the labeled masks

From the t-SNE plots shown in Fig. 2, we can also observe that (f) and (j), which represent the t-SNE plots of FR 4 and FR 8, have changed a lot. Compared to those of FR1, FR3, FR 5, and FR 7, the cluster of blue points move the positions from inside the package to outside. This is the reason the SR values of FR 4 and FR 8 increase significantly. This suggests that the feature representations generated through the PC maps of the concatenated feature maps can significantly improve the inter-cluster separation and intra-cluster compactness of the labeled pixels.

Fig. 3 shows the top 6 feature maps or PC maps corresponding to FR 1 to FR 8.

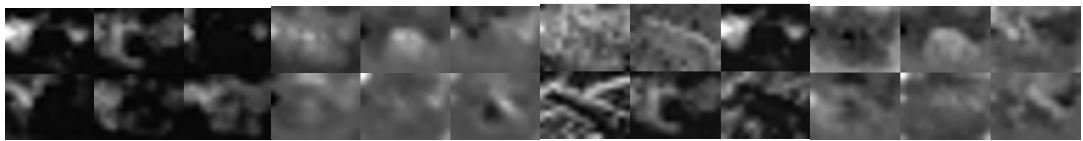

(a) FR 1              (b) FR 2              (c) FR 3              (d) FR 4

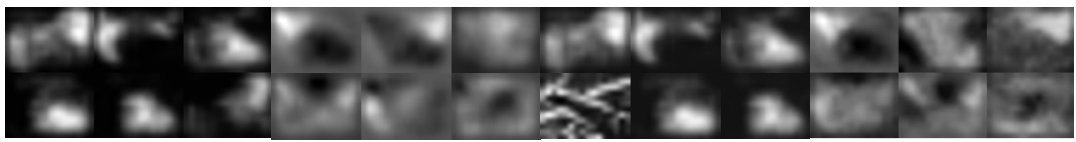

(e) FR 5              (f) FR 6              (g) FR 7              (h) FR 8

Fig. 3 Top 6 feature maps or PC maps of FR1 to FR 8

Based on the results shown in Fig. 3, FR 4 and FR 8 appear to capture the holistic features of the ingredients in the image.

Another issue from the observations is that using different backbones and different clustering methods can yield better segmentation for different images. Fig. 4 is an example regarding this issue.

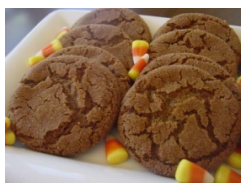

(a) Original image

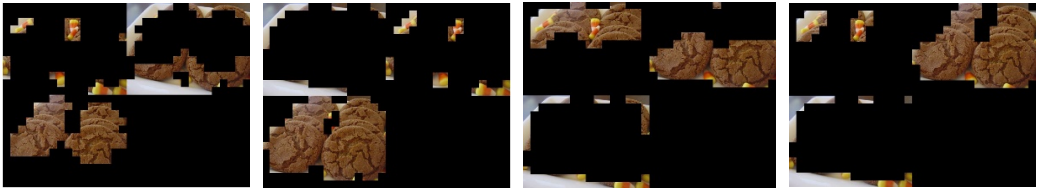

(b) K-means by FR 4 (c) Hierarchical by FR 4 (d) K-means by FR 8 (e) Hierarchical by FR 8

Fig. 4 Different segments from K-means clustering and hierarchical clustering

The original image is segmented into three parts. The results in (b) are from the k-means clustering method using the FR 4 derived from efficientNetB0; those in (c) are from the hierarchical clustering using the same FR 4; those in (d) are from the k-means clustering method using the FR 8 derived from pruned efficientNetB0; those of (e) are from the hierarchical clustering method using the same FR 8. From the segmentation results shown in Fig.4, we can see that the best segments are achieved in the case of (c).

Fig.5 shows the t-SNE plots of FR 4 and FR 8 corresponding to the labeled masks.

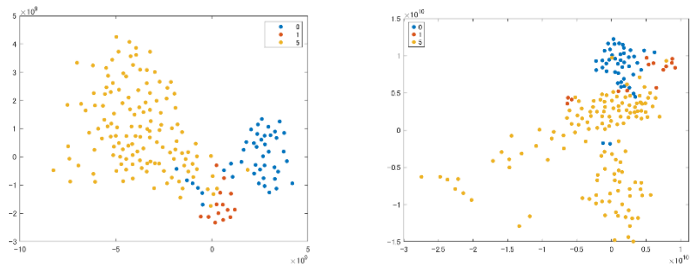

(a)  FR 4                                (b) FR 8

Fig. 5 An example of t-SNE plots of FR 4 and FR 8 corresponding to the labeled masks

Apparently, the t-SNE plot in (a) related to FR 4 exhibits better clustering quality in terms of inter-cluster separation and intra-cluster compactness. This is why the segments generated by FR 4 are superior to those from FR 8, regardless of whether k-means or hierarchical clustering is used. The compactness of the yellow points in (b) related to FR 8 is poor. This explains why the object (biscuit) is segmented into two parts when using k-means clustering, as shown in Fig. 4 (d). However, the biscuit is not segmented into two parts when using hierarchical clustering, as shown in Fig. 4 (e). This suggests that hierarchical clustering is more robust to issues with intra-cluster compactness.

## 4.  Approach for image segmentation

Base on the above investigation, we propose a novel architecture to enhance food image segmentation. An overview is provided in Fig. 6.

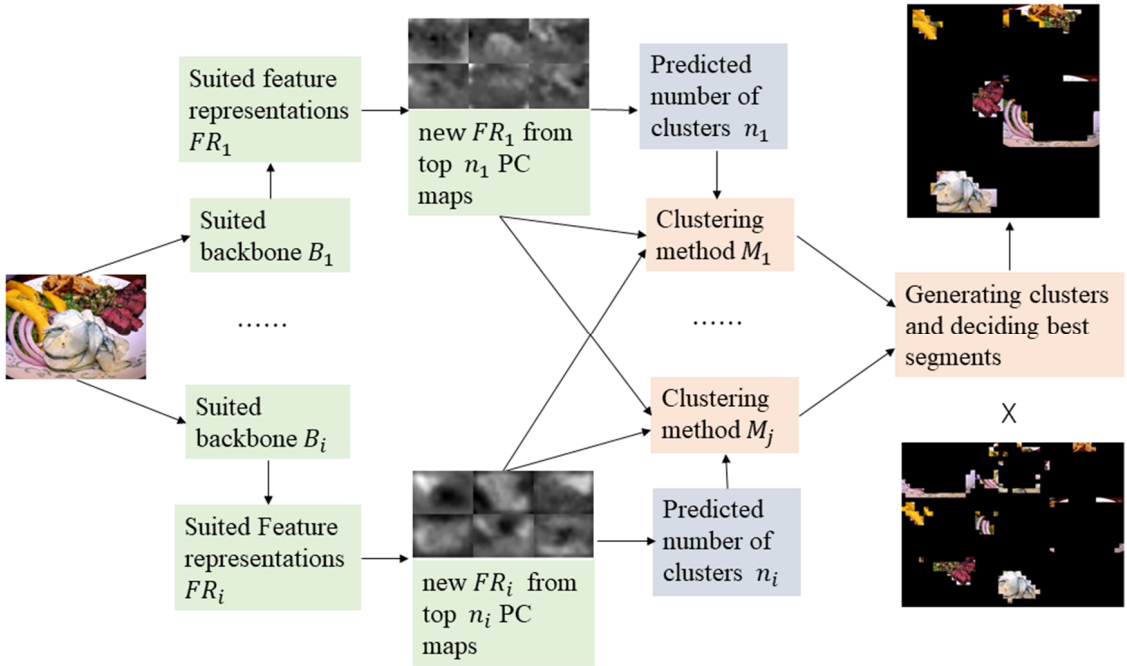

Fig. 6 An overview of proposed architecture

Based on the observation that ResNet and SAM do not serve as optimal backbones for providing appropriate feature representations for all images in terms of clustering quality, while EfficientNetB0 and pruned EfficientNetB0 perform better in this regard, we conduct a module for selecting suited backbones from several pre-trained models, including convolutional neural networks (CNNs). Based on the observation that pixel-wise feature representation array generated by concatenating feature maps from multiple convolutional layers lead to better inter-cluster separation and intra-cluster compactness for ingredients in images, we conduct a module for selecting appropriate feature representation arrays from several concatenated feature maps. Based on the observation that top PC maps of the pixel-wise feature presentation array can enhance the clustering quality for the ingredients in the images, we conduct a module for calculating PC maps and deciding the number of selected PC maps. Moreover, the number of selected PC maps is used as the number of clusters in the pixel-wise clustering, based on the consideration that these PC maps capture the distinctive features of each cluster. Based on the observation that using different backbones and different clustering methods can yield better segmentation for different images, we prepare several clustering methods, such as k-means and hierarchical clustering. The different feature presentations are fed to the various clustering models to get the different clustering results. Then, we introduce a metric to determine the best clusters, which are used to generate masks for image segmentation. In the following, we provide a detailed explanation of the modules and corresponding algorithms.

For selecting the appropriate backbone $B_i$ and feature representation array $FR_i$, the method proposed in [17] is employed. The metric SR, defined in (1), is used to select the most suitable $FR_i$ from those generated by different backbones and methods. According to the paper, it is optimal to select two or three of the most appropriate feature representation arrays derived from two of the most appropriate backbones for an image.

For obtaining new feature representation array from the PC maps of $FR_i$, an algorithm is proposed as shown in Fig. 7.

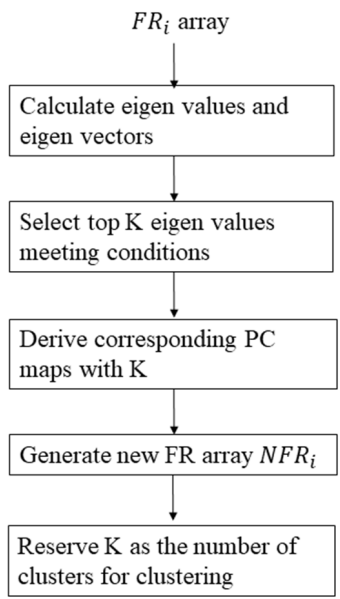

Fig. 7 Flowchart of generating new feature representation array

The eigenvalues and eigenvectors of pixel-level $FR_i$ array are calculated. The top K eigenvalues are selected, where the ratio of each eigenvalue to the maximal eigenvalue exceeds a threshold $T$. In this paper, $T$ is set to 0.3. Accordingly, based on the top K eigenvalues and their corresponding eigenvectors, the K principal component (PC) maps are derived. These PC maps form the new feature presentation array $NFR_i$. Furthermore, the value of K is used as the number of clusters for the clustering process in the next module.

The different pixel-wise $NFR_i$ arrays are clustered using various clustering methods, with the number of clusters set to K. In this paper, k-means clustering and hierarchical clustering methods are utilized. Based on the clustering results, a set of SRs corresponding to the formed clusters is calculated using equation (1). The clustering result with the highest SR value is then selected to generate a set of masks. These masks are used to segment the input image and obtain a set of segments.

Fig. 8 is an example showing the segments of an image using the proposed segmentation approach. Here, the pruned efficientNetB0 with removing some blocks is selected as the backbone, and the feature maps from head, block 15 and block 5 are concatenated to form the pixel-wise feature representation array. The hierarchical clustering method is utilized, and the number of clusters is set to 5 according the number of top eigenvalues that meet the condition that the ratio of each eigenvalue to the maximal eigenvalue exceeds 1/3.

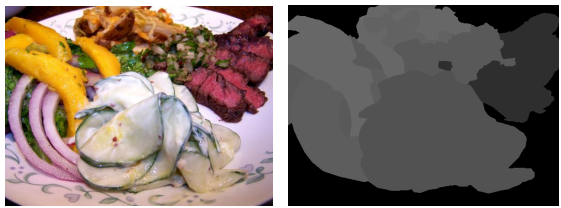

(a) Original image     (b) labeled masks

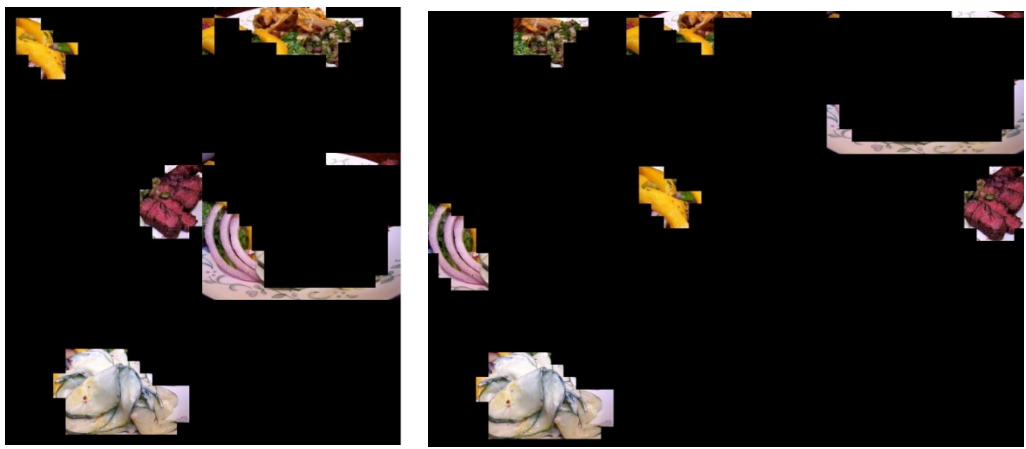

(c) Segments when K=5     (d) Segments when K=7

Fig. 8 masks and segments of an image

The five segments shown in (c) are the results when the number of clusters is set to 5, as predicted based on the number of top eigenvalues. The seven segments shown in (d) correspond to the results when the number of clusters is set to 7, matching the number of labeled masks. Comparing the segments in (c) to those in (d), we observe that although the predicted number of clusters does not match the ground truth (GT), reasonably accurate segmentation results can still be achieved.

## 5. Experiments and analysis

We conduct evaluation studies using MATLAB 2024b on a Dell laptop equipped with a 12GB RTX 4090 Laptop GPU and a 64GB RAM.

*4.1 Evaluation metrics*

We use pixel accuracy (pAcc) and mean Intersection over Union (mIoU) to evaluate the performance of image segmentation. Pixel Accuracy is a metric measuring the ratio of correctly predicted pixels to the total number of pixels in the image. This metric is particularly useful for understanding the overall accuracy of segmentation methods. For general pixel-wise accuracy without breaking it down per class, the mathematical expression for it is:

$$pAcc = \frac{Number\ of\ Correctly\ Classified\ Pixels}{Total\ number\ of\ pixels} \quad (2)$$

The IoU for a single class is the ratio of the intersection (overlap) of the predicted and ground truth areas to their union. For a multi-class segmentation task, mIoU is the average IoU across all classes.

The mathematical expression for mIoU is:

$$mIoU = \frac{1}{N} \sum_{i=1}^{N} \frac{TP_i}{TP_i + FP_i + FN_i} \quad (3)$$

where, $TP_i$ (true positives) is the number of correctly predicted pixels belonging to class $i$, $FP_i$ (false positives) is the number of pixels incorrectly predicted as class $i$ but belonging to another class, $FN_i$ (false negatives) is the number of pixels of class $i$ in the ground truth that were misclassified, and N is the total number of classes. A higher mIoU score indicates better segmentation accuracy, meaning that the predicted masks align more closely with the ground truth masks across all classes. In this paper, N is assigned as the number of clusters.

*4.2 Ablation studies*

We conduct ablation studies using 2,135 test images from the public dataset FoodSeg103 [18]. We firstly evaluate the performances of various backbones on image segmentation. Table 2 summarizes the average pAcc and the average mIoU of these images using the feature representations generated through the top K PC maps of the feature maps of the last convolutional layer from backbones of resNet50, SAM, effcientNetB0, and pruned efficientNetB0 with removing blocks of 6, 10, 13, and 14. Moreover, both k-means and hierarchical clustering methods are used in the image segmentation procedure, and the number of clusters for each image is set to K.

Table 2 Segmentation performance of various backbones

| backbone | pAcc | mIoU |
|---|---|---|
| resNet50 | 0.7448 | 0.3599 |
| SAM | 0.7514 | 0.3616 |
| efficientNetB0 | 0.8162 | 0.5095 |
| pruned efficientNetB0 | 0.7822 | 0.4976 |

Apparently, the pAcc and mIoU scores of efficientNetB0 and pruned efficientNetB0 are significantly higher than those of ResNet50 and SAM, particularly for mIoU. This is sufficient to demonstrate that, despite their widespread use as backbones, ResNet50 and SAM may not be the optimal choices for image segmentation. Moreover, although the pAcc and mIoU scores of efficientNetB0 are slightly higher than those of pruned efficientNetB0, pruned efficientNetB0 remains a strong choice as a backbone, given its advantages in memory efficiency and computation speed.

Next, we evaluate the performance of various feature representations from EfficientNetB0 (EffB0) and pruned EfficientNetB0 (EffB0-P) using different clustering methods for image segmentation. In this paper, k-means and hierarchical clustering method are utilized. Table 3 summarizes the corresponding average pAcc and the average mIoU of the test images. Here, "Abl." stands for "ablation study" "Cat." refers to the concatenation of feature maps, "PC" refers to the principle component maps, "Clus." refers to the clustering method, "o" denotes "adopting", and "x" denotes "not adopting". Specifically, "o" under "Cat." indicates that the feature maps of the head, block 15, and block 5 of EfficientNetB0 are concatenated, while "x" indicates that only the feature maps of the head are used. Moreover, "both" under "backbone" indicates that both EfficientNetB0 and pruned EfficientNetB0 are used in the image segmentation procedure, while "both" under "Clus." indicates that both k-means and hierarchical clustering methods are utilized. We omit cases that do not use PC maps, as their gAcc and mIoU scores are lower than those of cases that do. Additionally, we omit cases where only a single clustering method is applied for the same reason.

Table 3 segmentation performance of various feature presentations

| Abl. | backbone | Cat. | PC | Clus. | pAcc | mIoU |
|---|---|---|---|---|---|---|
| A1 | effB0 | x | o | both | 0.8162 | 0.5095 |
| A2 | effB0 | o | o | both | 0.8454 | 0.5214 |
| A3 | effB0-P | x | o | both | 0.7822 | 0.4976 |
| A4 | effB0-P | o | o | both | 0.7957 | 0.5320 |
| proposed | both | o | o | both | 0.8308 | 0.5423 |

For the ablation studies A1, A2, A3, and A4, the results in Table 3 show that both pAcc and mIoU scores increase when feature map concatenations are applied. For EfficientNetB0, pAcc improves by approximately 0.03, and for pruned EfficientNetB0, mIoU increases by about 0.03. As for proposed approach, the mIoU score of the proposed method is highest at 0.54, exceeding the second-highest by 0.01, and the pAcc score is second-highest at 0.83, which is 0.015 lower than the first place. These results align with the observations discussed in Section 2, demonstrating that the proposed segmentation method is both rational and effective. Furthermore, it should be noted that case A4, where a pruned EfficientNetB0 is used as the backbone, is suitable when prioritizing memory efficiency and computation speed.

*4.3 Comparison to SOTA research*

Furthermore, we compare the segmentation performance to SOTA results. The comparisons are made on FoodSeg103 [18] and UECFOODPIX COMPLETE [24] datasets which are widely used in evaluating the performance of food image segmentation. Table 4 summarizes the results of SOTA research and ours on FoodSeg103 dataset. "Proposed (K: predicted)" refers to our proposed method, where the number $K$ of clusters is predicted based on the selected eigenvalues. "Proposed (K: GT)" indicates that the number of clusters corresponds to the number of ground truth (GT) masks.

Table 4 Comparison to SOTA results on FoodSeg103 dataset

| Model | pAcc | mIoU |
|---|---|---|
| CCNet-Finetune[18] | 0.8770 | 0.4130 |
| FoodSAM [20] | 0.8410 | 0.4642 |
| BEiTv2Large [21] |  | 0.494 |
| OVFoodSeg [22] |  | 0.381 |
| Proposed (K: predicted) | **0.8308** | **0.5423** |
| Proposed (K: GT) | 0.8417 | 0.5415 |

As shown in Table 4, our method outperforms the SOTA approaches, achieving a mIoU score of 0.54, an improvement of approximately 0.05 over the second-highest SOTA score. Moreover, it is noticed that the mIoU score of the Proposed (K: predicted) is almost same as that of Proposed (K: GT). This demonstrates that using the number of selected eigenvalues as the number of clusters for segmenting ingredients is both rational and effective. It should be noted that the proposed method performs class-agnostic segmentation, whereas the SOTA approaches listed in Table 3 are class-based segmentation methods. Accordingly, we further evaluate the recognition performance of the obtained segments. Using the method proposed in [16] combined with a pruned EfficientNet-B0 model fine-tuned on SI110 with 110 ingredient categories, we assign segments to their corresponding ingredients. Since FoodSeg103 and SI110 share only 38 identical ingredient categories, we evaluate the image-wise precision, recall, and F1 score approximately for these 38 categories. For the test samples in FoodSeg103, the average precision, recall, and F1 scores are 0.4251, 0.4133, and 0.4612, respectively.

While the results are promising, the model faces challenges in accurately classifying segments with high inter-class similarity. Improving the ingredient classification model to better handle high inter-class similarity could lead to higher recognition performance for the segments.

Table 5 summarizes the results of SOTA research and ours on UECFOODPIX COMPLETE dataset.

Table 5 Comparison to SOTA results on UECFOODPIX COMPLETE dataset

| Model | pAcc | mIoU |
|---|---|---|
| BayesianGourmetNet [19] | 0.8805 | 0.6616 |
| FoodSAM [20] | 0.8847 | 0.6614 |
| Proposed (K: predicted) | **0.9097** | **0.4939** |
| Proposed (K: GT) | 0.8840 | 0.6422 |

As shown in Table 5, the mIoU score of the proposed method with the predicted cluster number is about 0.16 lower than the SOTA results, although the pAcc is slightly higher. However, it is observed that the mIoU score of the proposed method, when the cluster number is set to the number of labeled masks, is nearly the same as the SOTA results. This is because the masks of this dataset are labeled confusedly with either dish names or ingredient names, unlike FoodSeg103, where the masks are labeled exclusively with ingredient names. Accordingly, the number of clusters cannot be accurately predicted based on the eigenvalues, which degrades the performance of ingredient segmentation. The fact that the mIoU reaches the SOTA level when the number of clusters matches the number of masks demonstrates, from a contrasting perspective, that using the number of selected eigenvalues as the cluster count for ingredient segmentation is both effective and rational. This further indicates that the segmentation results are independent of the datasets, revealing the potential of the proposed method in subsequent open-ingredient classification.

### 4.4 Examples of segmentation

Some segmentation results of segmenting ingredients from food images using the proposed approach are shown in Fig, 9. There are examples of variability in ingredient appearance, occlusion of ingredients, and complex and clutter backgrounds Among these images.

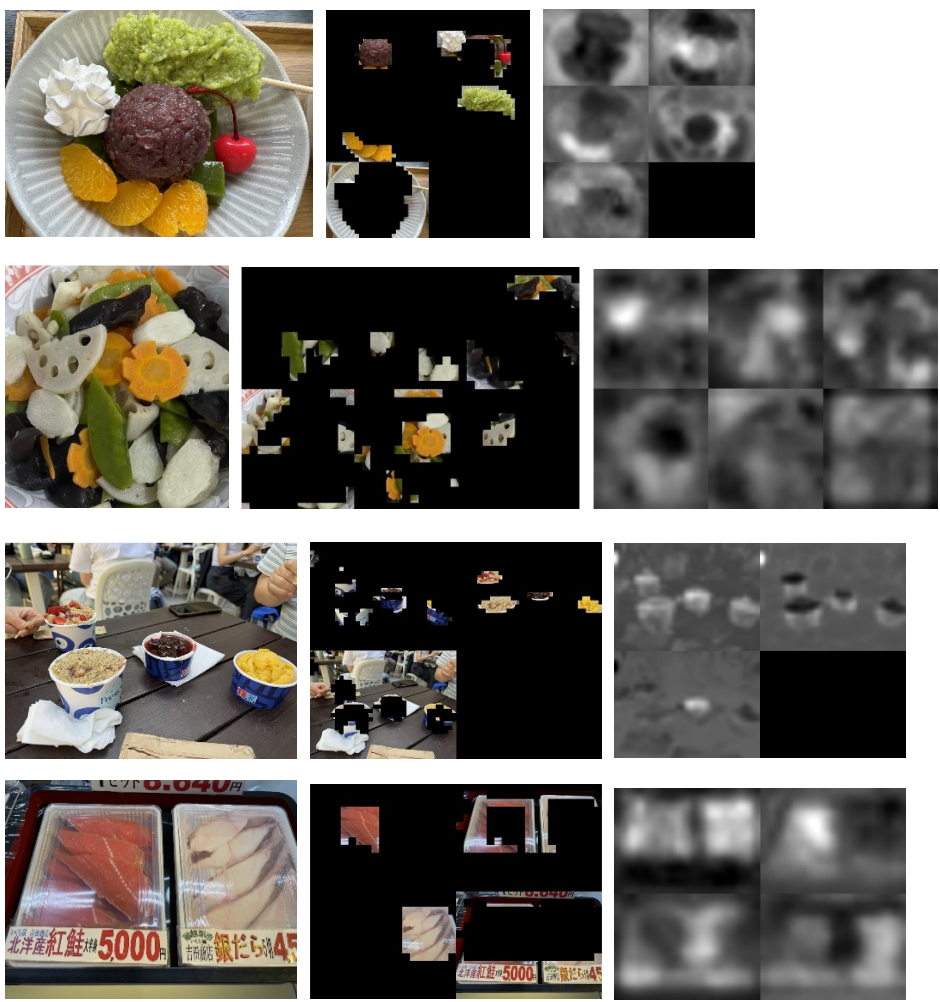

Fig. 9 Examples of segmentation

The images in the first column of Fig. 9 are the original images. The images in the second column are segmented versions of these originals, and those in the third column represent the top *K* selected PC maps, based on the eigenvalues of the feature representations. Note that the number of segments corresponds to the number of selected PC maps. As shown in Fig. 9, the ingredients in the images are well-separated despite variability in ingredient appearance, occlusion of ingredients, and complex and clutter backgrounds.

6. **Distinct discussion**

Our method is proposed based on the experimental observations and validated using the public datasets. This indicates that our method offers greater interpretability than existing SOTA methods which are based on the black-box methods like multi-label learning or semantic segmentation. The proposed segmentation method is class-agnostic and has potential in subsequent open-ingredient classification. This means it can offer greater flexibility, generalization, and interpretability while reducing the need for exhaustive labeled datasets. This approach is increasingly important in real-world applications where diverse and dynamic environments require adaptive and robust segmentation solutions, such as in agricultural product monitoring and food production.

7. **Conclusion**

For open vocabulary recognition of ingredients in food images, segmenting the ingredients is a crucial step. This paper proposed a novel approach that explores PCA-based feature representations of image pixels via CNNs to enhance segmentation. Empirical experiments were conducted to examine the impact of different backbones and feature representations on pixel clustering quality. To evaluate this, an internal clustering metric based on the silhouette score was defined. Using this metric, we explored the selection of optimal feature representations and suitable clustering methods for ingredient segmentation.

Additionally, it was observed that principal component (PC) maps derived from concatenated backbone feature maps improved the clustering quality of pixel-wise feature representation array, resulting in more stable segmentation outcomes. A new method for predicting the number of clusters was also introduced, based on the eigenvalues of the feature representation array. Specifically, using the number of selected eigenvalues as the number of clusters yielded effective segmentation results.

Based on these experimental observations, we proposed and implemented a new architecture for segmenting multiple ingredients in food images. Experimental results showed that the proposed method outperformed the SOTA results on the ingredient-labeled dataset FoodSeg103, achieving a mIoU score of 0.5423, when both EfficientNetB0 and pruned EfficientNetB0 are used for deriving pixel-level feature representation, and both k-means and hierarchical clustering methods are utilized for segmenting ingredients. Notably, the proposed method is unsupervised, with pixel-level feature representations from backbones not fine-tuned on specific datasets. This highlights the flexibility, generalizability, and interpretability of the proposed method, while reducing the need for extensive labeled datasets.

In future work, it is necessary to develop a framework that supports an open-ingredient setting to address the challenges of open-set ingredient classification.

**Acknowledgments**

This work was partly supported by JSPS KAKENHI Grant Number JP22K12095 and JKA Subsidy Program for Keirin and Auto Racing.